\documentclass[conference]{IEEEtran}
\IEEEoverridecommandlockouts
\usepackage{cite}
\usepackage{amsmath,amssymb,amsfonts}
\usepackage{algorithmic}
\usepackage{graphicx}
\usepackage{textcomp}
\usepackage{xcolor}
\def\BibTeX{{\rm B\kern-.05em{\sc i\kern-.025em b}\kern-.08em
    T\kern-.1667em\lower.7ex\hbox{E}\kern-.125emX}}
\begin{document}

\title{Comparison between Docker and Kubernetes based Edge Architectures for Enabling Remote Model Predictive Control for Aerial Robots}
\author{Achilleas Santi Seisa, Sumeet Gajanan Satpute and George Nikolakopoulos%
\thanks{This work has been partially funded by the European Unions Horizon 2020 Research and Innovation Programme AERO-TRAIN under the Grant Agreement No. 953454.}
\thanks{The authors are with the Robotics and AI Team, Department of Computer, Electrical and Space Engineering, Lule\aa\,\, University of Technology, Lule\aa\,\,}
\thanks{Corresponding Author's email: {\tt\small achsei@ltu.se}}
}
\maketitle
\begin{abstract}
Edge computing is becoming more and more popular among researchers who seek to take advantage of the edge resources and the minimal time delays, in order to run their robotic applications more efficiently. Recently, many edge architectures have been proposed, each of them having their advantages and disadvantages, depending on each application. In this work, we present two different edge architectures for controlling the trajectory of an Unmanned Aerial Vehicle (UAV). The first architecture is based on docker containers and the second one is based on kubernetes, while the main framework for operating the robot is the Robotic Operating System (ROS). The efficiency of the overall proposed scheme is being evaluated through extended simulations for comparing the two architectures and the overall results obtained.
\end{abstract}

\section{Introduction}
\label{intro}
Nowadays, the need for autonomous solutions for robotic applications is rapidly increasing, while these operations are usually based on complex methodologies and algorithms, which are computationally demanding. In many cases, robotic platforms can not handle the demanding processes since they do not have the needed computational power. In that context, it is crucial to increase the computational capability of robots, which is one of the main reasons why the idea of utilizing external resources for robotic applications has been studied extensively. 
Some missions, like controlling and navigating a robot, are delay sensitive and thus, cloud solutions can not be used for these kind of applications and instead, edge computing solutions, in a form of distributed computing can be used. Towards this direction, related works in the field of cloud architectures for vehicles, the different layers and the subcategories are presented in~\cite{9184917}. Edge computing not only provides significant improvements in resources in terms of processors that the robots are deployed with, but also, the issue of reduced time delays, since the hardware of edge computing is located in a close proximity to the robots. As a result, the time delays for the data to be transferred from the robots to the edge and vice versa, are relatively small. Even though edge computing has been proven to be a promising solution, it is still not universally applied especially for closed-loop systems. 

Virtual Machines (VMs) are a good option for heavy applications since they emulate an entire machine, down to the hardware layers, but not for application for which mobility needs to be supported. For these applications docker containers~\cite{7931566}, which emulate only the software components, are a better option. In~\cite{7300842}, a review of containers and clusters with their benefits compared to VMs are presented. Researchers also discussed the edge cluster requirements and the option of containerized application orchestrators like kubernetes. Docker and kubernetes are two very popular platforms among researchers and engineers, because they allow them to build and deploy distributed applications quickly. These technologies have been used for cloud and edge based robotic applications. In~\cite{8651759}, a containerized application including a remote controller that could run through a mobile edge server in a form of a docker container is introduced. An optimisation strategy, based on stochastic processes is introduced in~\cite{kochovski2019architecture} for distributing containers to the cloud, edge and fog, while the whole process is automated by using a kubernetes orchestration. 
An architecture for distributed edge and cloud resources is presented in~\cite{figueiredo2020edgevpn}
based on kubernetes orchestration for a virtual cluster, while in~\cite{cha2021design}, low-latency edge services for robotic applications were introduced for container interaction. In~\cite{lumpp2021container}, an architecture based on docker, kubernetes and ROS, for robotic applications is presented. The applications were deployed into the cloud clusters, and the architecture was evaluated through several configurations of the interaction of a mobile robot with an industrial agile production chain. 

In~\cite{skarin2020cloud2} and~\cite{aarzen2018control}, researchers suggested to offload the model predictive controller (MPC) on external sources. A cloud-based architecture was proposed in~\cite{skarin2020cloud2} to offloaded the MPC with a variable horizon strategy. 
In~\cite{aarzen2018control}, an architecture consisting of two MPCs is implemented where one MPC is running on a local edge and the other on a cloud. In comparison to these articles, our current work focuses more on the architecture of the system, while investigating different frameworks and components.

The main contributions of this article stems from presenting and highlighting two potential architectures that can be used, not only for controlling the trajectory of an aerial robot, but also to be used for closing the loop between the edge and any platform for time sensitive missions. The proposed solutions are based on technologies that most edge providers are offering, which makes these solutions universally applicable. Our goal is to point out the main components and the operating principles of these architecture, and showcase their pros and cons through simulations. That way, we are aiming to present the advantages of applied edge computing in the field of robotics and more specific to aerial robotics, and at the same time, provide inspiration for future works.

The rest of the article unfolds in the following manner. In Section~\ref{mpc} the model of the UAV and the controller are presented. In Section~\ref{architecture} the two different architectures are described and the technologies used for each one of them are presented. Simulation results are demonstrated and meaningful characteristic of the architectures, such as responses and time delays are shown in Section~\ref{simulation}. Finally, we conclude the article in Section~\ref{conclusion}, with a brief overview and potential future directions.

\section{Model Predictive Control}
\label{mpc}
The utilized controller for this article is a Model Predictive Controller, based on~\cite{9143931}. MPC is a commonly used controller for aerial robots, thanks to its advantages and predictive behaviour. In comparison to other commonly used controllers such as PID or LQR, MPC is a more computationally heavy controller, since it uses optimization methods to extract the future states based on a finite prediction horizon, thus, external resources can be handy. 

The UAV is considered as a six degree of freedom robot with a fixed body frame and the MPC is responsible for controlling the trajectory of the UAV. The kinematic model is described by Eq.~\ref{eq:kinematics} in the body frame.

\begin{align}
&\dot{p}(t) = v_{z}(t) \nonumber\\
&\dot{v}(t) = R_{x,y}(\theta,\phi) \begin{bmatrix} 0\\ 0\\ T\end{bmatrix} + \begin{bmatrix} 0\\ 0\\ -g\end{bmatrix} - \begin{bmatrix} A_{x} & 0 & 0\\ 0 & A_{y} & 0\\ 0 & 0 & A_{z}\end{bmatrix}u(t) \label{eq:kinematics}\\
&\dot{\phi}(t) = \frac{1}{\tau_{\phi}} (K_{\phi} \phi_{ref}(t) - \phi(t)) \nonumber\\
&\dot{\theta}(t) = \frac{1}{\tau_{\theta}} (K_{\theta} \theta_{ref}(t) - \theta(t)) \nonumber
\end{align} 

In Eq.~\ref{eq:kinematics} we denote the following parameters. $p$ is the position, and $v$ is the linear velocity given in the global frame. $R(\phi(t), \theta(t))$ is the rotation matrix that represents the attitude in Euler form. $\phi$ and $\theta$ are the roll and pitch angles along the $x^{\mathbb{W}}$ and $y^{\mathbb{W}}$ axes respectively, while $\phi_{d}$ and $\theta_{d}$ and $T \geq 0$ are the desired inputs values to the system in roll, pitch and the total thrust. In this model, the acceleration is depending only on the magnitude and angle of the thrust vector, produced by the motors, as well as the linear damping terms symbolized as $A_{x}, A_{y}, A_{z}$ and the gravity of earth symbolized as $g$. The attitude terms are modeled as a first-order system between the attitude and the referenced $\phi_{ref}$ and $\theta_{ref}$, with gains $K_{\phi}$ and $K_{\theta}$ and time constants $\tau_{\phi}$ and $\tau_{\theta}$. The motor commands for the UAV are generated through a lower-level attitude controller that takes thrust, roll and pitch commands as inputs in order to produce the commands.
\subsection{Cost Function}
\label{cost_function}
For the cost function, the state vector of the UAV is denoted as $x = [p, v, \phi, \theta]^{T}$ and the control input vector as $u = [T, \phi_{d}, \theta_{d}]^{T}$. The sampling time of the system is $\delta_{t}$, using a forward Euler method for each time instance $(k+1|k)$. The prediction considers the specified number of steps into the future, which is called prediction horizon and it is represented as $N$. A related optimizer is tasked with finding an optimal set of control actions, defined by the cost minimum of this cost function, by associating a cost to a configuration of states and inputs at the current time and in the prediction. The predicted states at the time step $k+j$, produced at the time step $k$ are represented as $x_{k+j|k}$. The corresponding control actions are represented as $u_{k+j|k}$. Also $x_{k}$ and $u_{k}$ represent the full predicted states and the corresponding control inputs along the prediction horizon correspondingly. The objective of the controller is to navigate to the desired position and deliver smooth control inputs. The cost function is presented in Eq.~\ref{eq:cost_funtion} as:

\begin{align}
&J = \sum_{j=1}^{N} \underbrace{(x_{d} - x_{k+j|k})^{T} Q_{x} (x_{d} - x_{k+j|k})}_{state \quad cost} \nonumber\\
&+ \underbrace{(u_{d} - u_{k+j|k})^{T} Q_{u} (u_{d} - u_{k+j|k})}_{input \quad cost} \label{eq:cost_funtion}\\
&+ \underbrace{u_{k+j|k} - u_{k+j-1|k})^{T} Q_{\delta u} (u_{k+j|k} - u_{k+j-1|k})}_{control \quad actions \quad smoothness \quad cost} \nonumber
\end{align} 
where $Q_{x}$ is the matrix for the state weights, $Q_{u}$ is the matrix for the input weights and $Q_{\delta u}$ is the matrix for the input rate weights. The first term describes the state cost, which is the cost associated with deviating from a certain desired state $x_{d}$. The second term describes the input cost that penalizes a deviation from the steady-state input $u_{d} = [g, 0, 0]$ and represent the inputs that describe hovering. The final term is added to guarantee that the control actions are smooth. 
We evaluate the overall behavior of the MPC scheme by measuring the overall time delays for the proposed edge architectures.

\section{Architecture and Frameworks}
\label{architecture}
The proposed closed loop systems consist of the edge side and the robot side. On the robot side we have the UAV that we are aiming to control and on the edge side we have the controller. The edge side of the system is different since we used two different edge solutions. In the first case, we used docker containers, which will be described in Section~\ref{docker} and in the second case, we used kubernetes orchestration which will be described in Section~\ref{kubernetes}. In both architectures, the utilized docker images and the robot side is the same. ROS is used for the operation of the robot and the communication between the edge and the robot through ROS nodes. Furthermore, we have to export ROS master and IP in every device so that communication between edge and UAV ROS nodes would be established.

In this work, all the devices are operating on the same network. This makes the setup easier to handle since ROS communication is based on opening random ports for the communication between ROS nodes of different devices. The described communication is enabled through a WiFi network.

The states $x(k)$, which are the linear position, linear velocity and quaternions, are generated by the robot and are published to the odometry ROS topic. The MPC ROS node subscribes to that topic, as well to the reference ROS topic, which publishes the referenced trajectory signal of the UAV, $r(k)$. The states signal arrives to the MPC ROS node with a time delay due to the travel time between the robot and the edge, thus the states signal on the MPC is $x(k-d_{1})$. The generated by the MPC command signal is denoted as $u(k-d_{2}$ and is published to the command topic to which the robot ROS node subscribes in order to receive that command signal. The command signal arrives to the robot with some delay so it is stated as $u(k-d_{3})$. Finally, the output of the system is represented as $y(k)$.

\subsection{Docker}
\label{docker}
\begin{figure}[ht!]
	\centering
	\includegraphics[width=0.95\columnwidth]{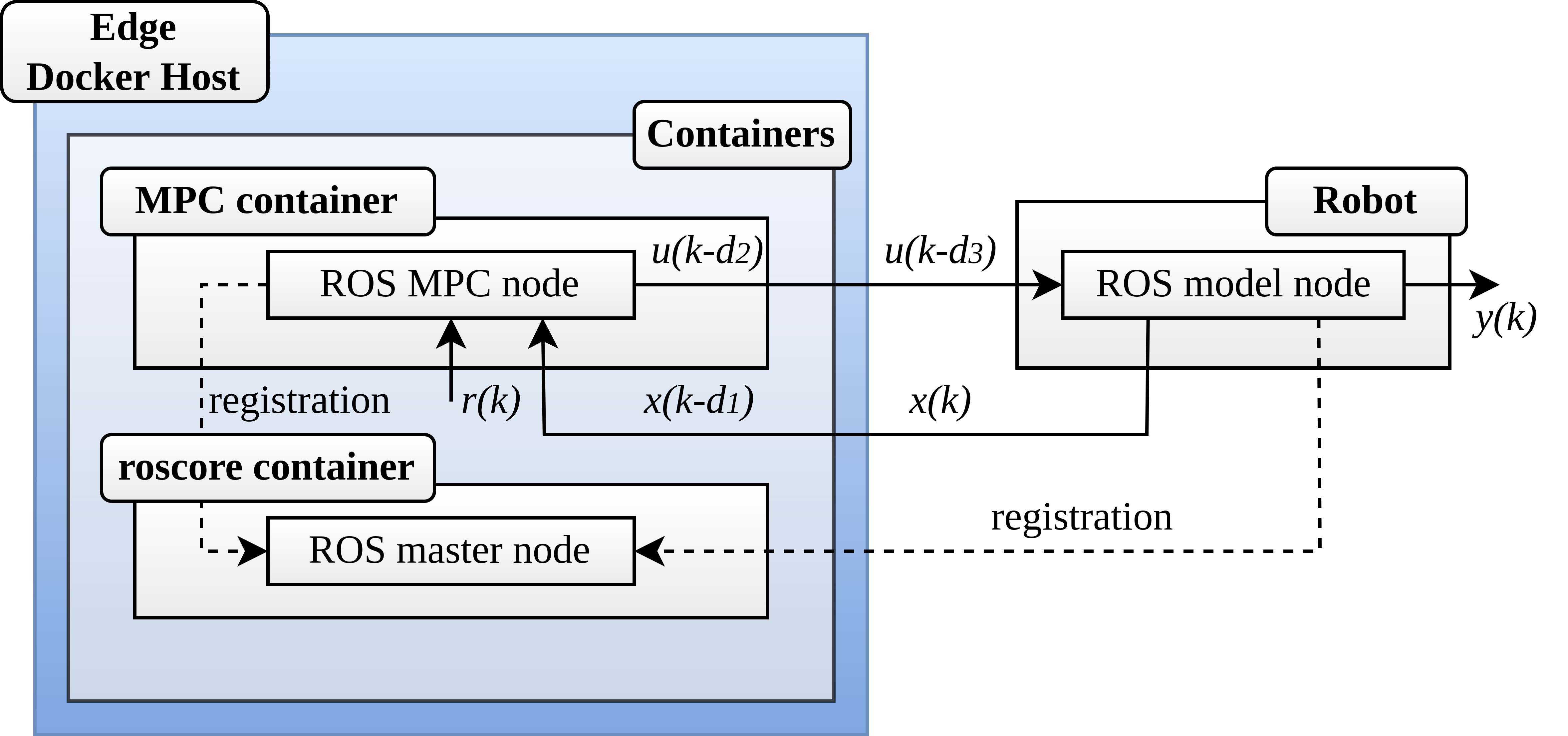}
  	\caption{Docker-based edge architecture. Data flow across ROS nodes is depicted for the closed loop system}
  	\label{fig:docker}
\end{figure}

Docker is a platform for developing and running and managing containerized applications quickly. For the docker-based edge architecture, we used two docker images that were deployed with all the necessary libraries and dependencies, in order to run the controller. Both docker images used ROS noetic on Ubuntu 20.04, entrypoint. On top of it, for one of the images, additional ROS packages, like the MPC package, optimization engines and some needed libraries were deployed. Afterwards, the two docker containers were created on an edge machine. The first docker container is running the ROS master, where all the ROS nodes have to register in order to communicate with each other, and the second docker container contains the MPC application and is running the controller. The docker architecture is shown in Fig.\ref{fig:docker}.

The communication between ROS nodes is through IP addresses and ports. Each device has its own IP in the network, while the assignment of ports is happening randomly. On the other hand, containerized applications are assigned to private IPs. This creates a communication issue among the containerized ROS nodes. In order to overcome this issue, the two containers are launched with access to the networking interfaces of the host by using the host network option when running the containers.

\subsection{Kubernetes}
\label{kubernetes}
\begin{figure}[ht!]
	\centering
	\includegraphics[width=0.95\columnwidth]{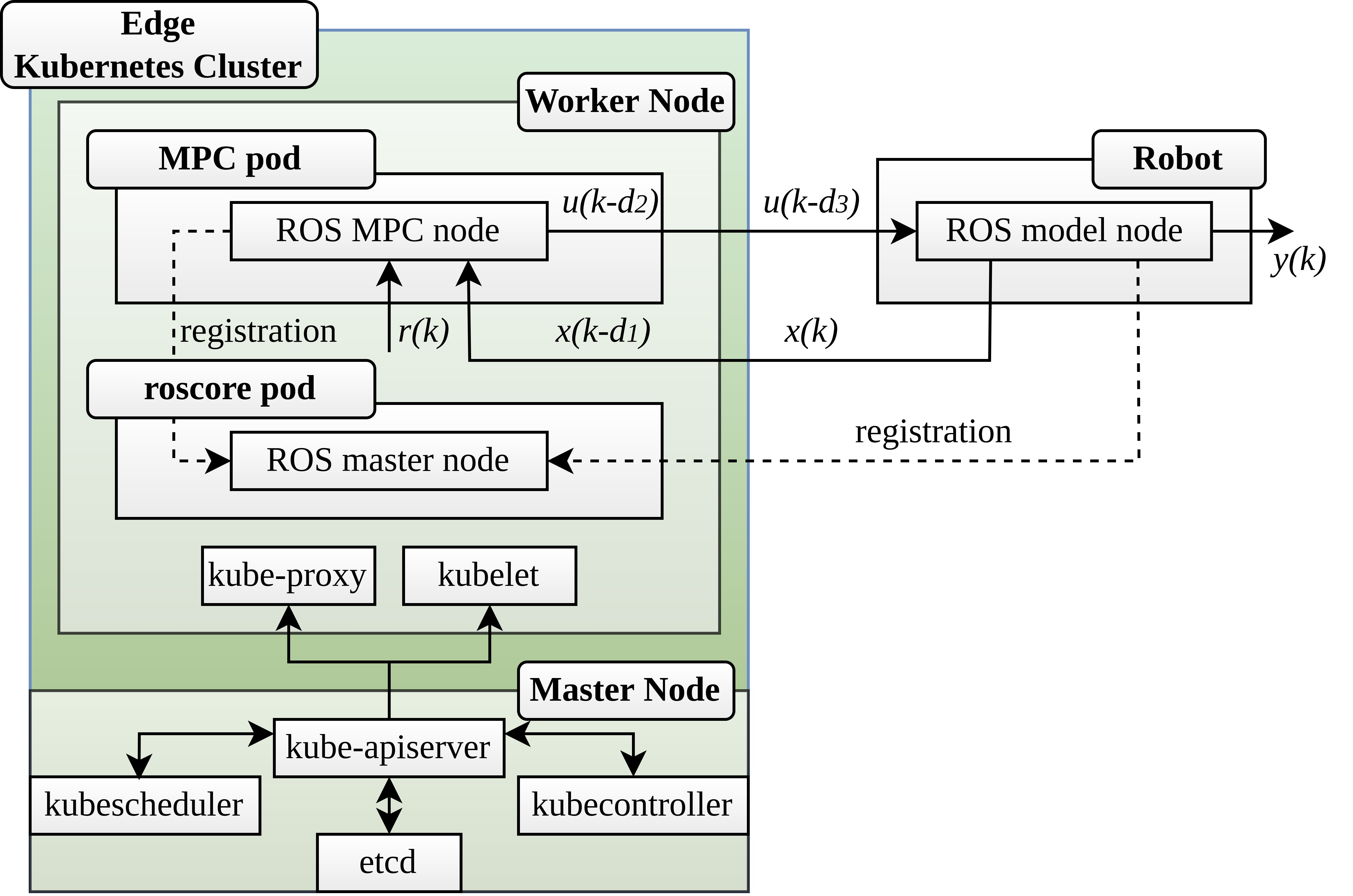}
  	\caption{Kubernetes-based edge architecture. Data flow across ROS nodes is depicted for the closed loop system}
  	\label{fig:kubernetes}
\end{figure}

Kubernetes is a containerized application orchestrator, mainly based on docker containers, and it offer several features. These features can be essential for some applications, but come with a cost since kubernetes is a much more complex environment in comparison to just using docker containers, and it consists of more components and requires some expertise. 
In the case of controlling a UAV through the edge, the features that kubernetes can provide, like the automated deployment of the application, scheduling, scaling and monitoring are essential and can create a more robust and resilient ecosystem for our application. A Kubernetes cluster consists of two nodes, the master node that controls and manages the worker node and the worker node where the application pods are deployed. The deployed pods are based on the same images that were created for the docker-based edge architecture. The first pod consists of the ROS master that is necessary for the ROS operation and the second pod consists of the controller and all the necessary libraries and dependencies. The kubernetes architecture is shown in Fig.\ref{fig:kubernetes}.

Kubernetes is a multi layer system and again in this case containerized applications are assigned to private subnet IPs. To overcome this problem, we used the host network option when we deployed the pods in order to use the ROS framework and give to the pods access to the loopback device. Additionally, we use services to enable the communication between the pods.

\subsection{Docker and Kubernetes Comparison}
\label{comparison}
In the above architectures, we used docker containers both independently and as a component of kubernetes cluster. The docker-based architecture was about packaging the MPC and roscore containerized applications on a single node, while the kubernetes-based was meant to run them across a cluster. For small projects, without a lot of requirements adopting docker might be efficient, but for larger projects with a lot of requirements, such as container scheduler, kubernetes can be much more beneficial. In our application, both architectures were efficient. However, if we want to expand our application for more complex missions and we need to deploy multiple containers, kubernetes provides some features, that can be essential. For example, re-creation is an important feature, even in our application, since in case of failure of the MPC pod, kubernetes will take care of it and will automatically deploy a new one.


\section{Simulation Results}
\label{simulation}
For the evaluation of the two architectures, we ran a series of simulations. To simulate the UAV, we used the simulation environment gazebo and the ROS package rotors simulator, on a local computer. For the edge, we had a machine with the following characteristics: Processor: Intel Core i5-8400 CPU at 2.80GHz×6 and RAM: 32GB. This machine was the host for both the docker containers and the kubernetes cluster. The two devices communicate with each other through ROS over the same WiFi network. 

The two presented architectures were evaluated in terms of time delays. To achieve that we measured the travel times and execution times, while the MPC execution rate was set at $100 Hz$ and the MPC horizon at $100 steps$. These values are relatively high, thus some local processor would not be able to handle the operation of the MPC. The advantage of having such high values is that the execution of the MPC is fast, which means that it can generate commands to be send to the UAV rapidly. Moreover, MPC can predict the change of dependent variables, thus high prediction horizon can make better predictions and predict changes faster. Since we are utilizing a powerful edge machine we were able to choose high values without any issue.


The first set of experiments is based on the docker architecture. We present the responses of the system for the three trajectories, we showcase the time measurements for the travel times and the execution time and we present the CPU usage.


\begin{figure}[ht!]
	\centering
	\includegraphics[width=0.95\columnwidth]{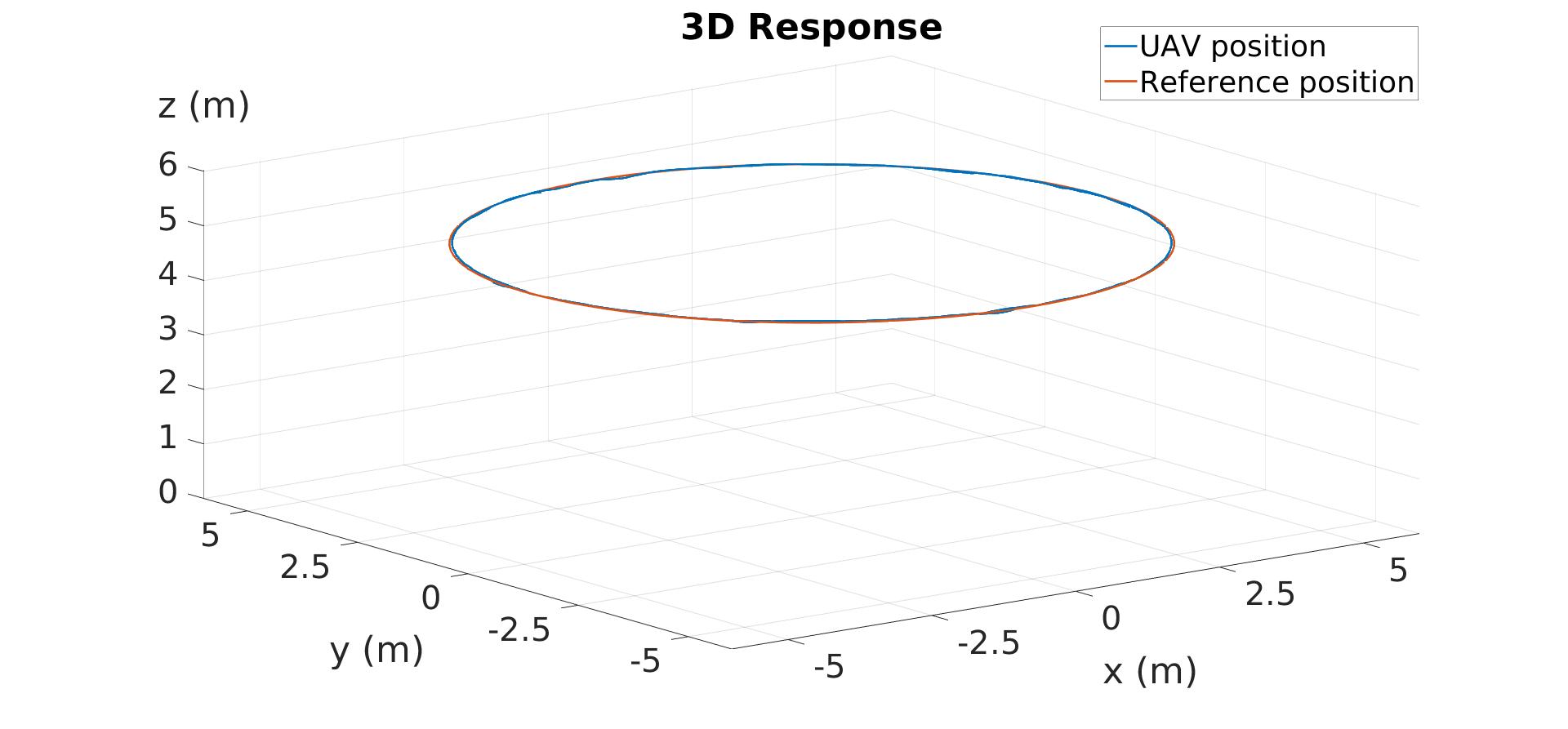}
  	\caption{UAV 3D circular trajectory based on a docker-based architecture}
  	\label{fig:dcircular3d}
\end{figure}

\begin{figure}[ht!]
	\centering
	\includegraphics[width=0.95\columnwidth]{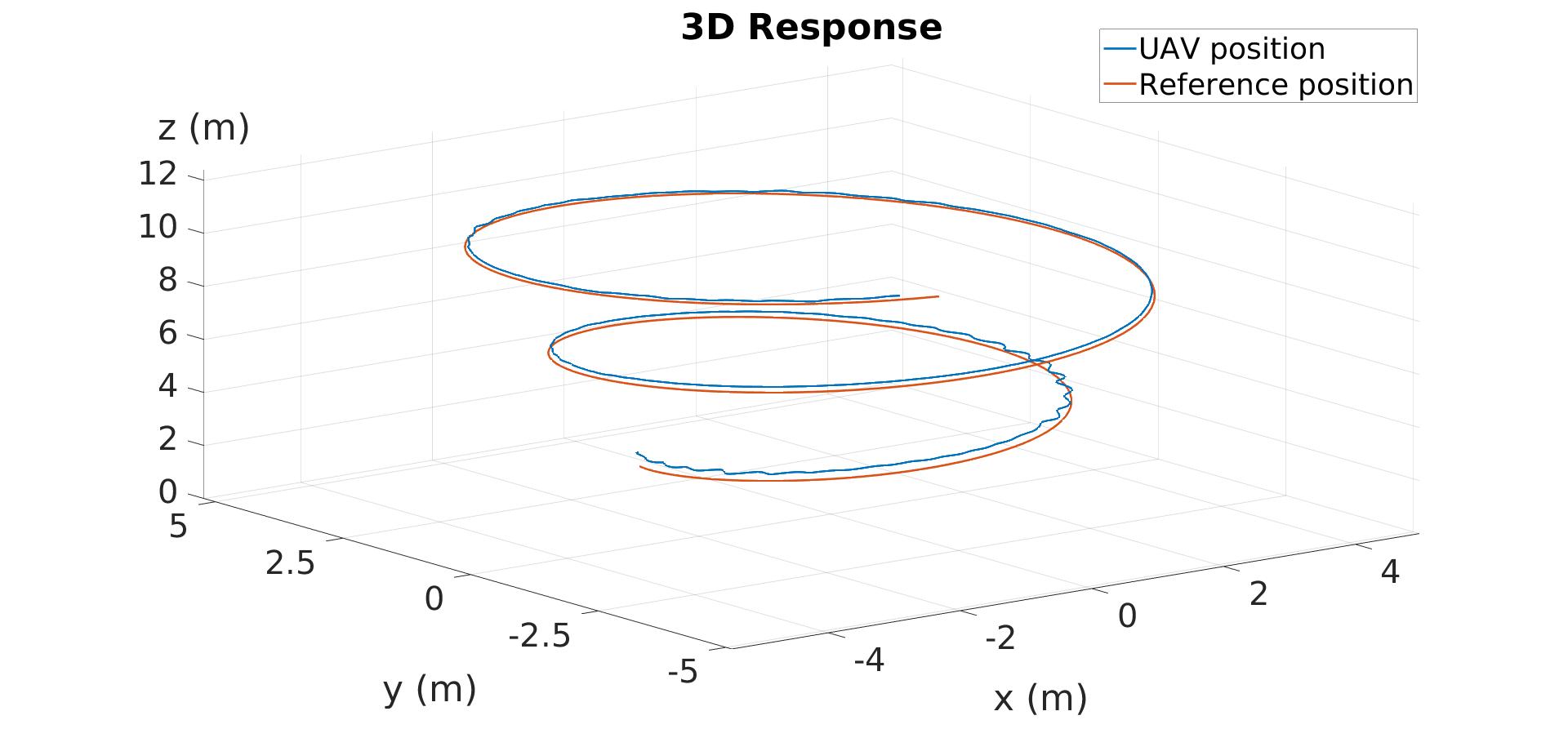}
  	\caption{UAV 3D helical trajectory based on a docker-based architecture}
  	\label{fig:dhelical3d}
\end{figure}

In~Fig.~\ref{fig:dcircular3d} and Fig.~\ref{fig:dhelical3d}, we can observe that the UAV (blue line) can follow the referenced trajectory (red line) in the desired manner. The offloaded MPC can control the UAV without any major issue according to the simulation results and the graphs.

\begin{figure}[ht!]
	\centering
	\includegraphics[width=0.95\columnwidth]{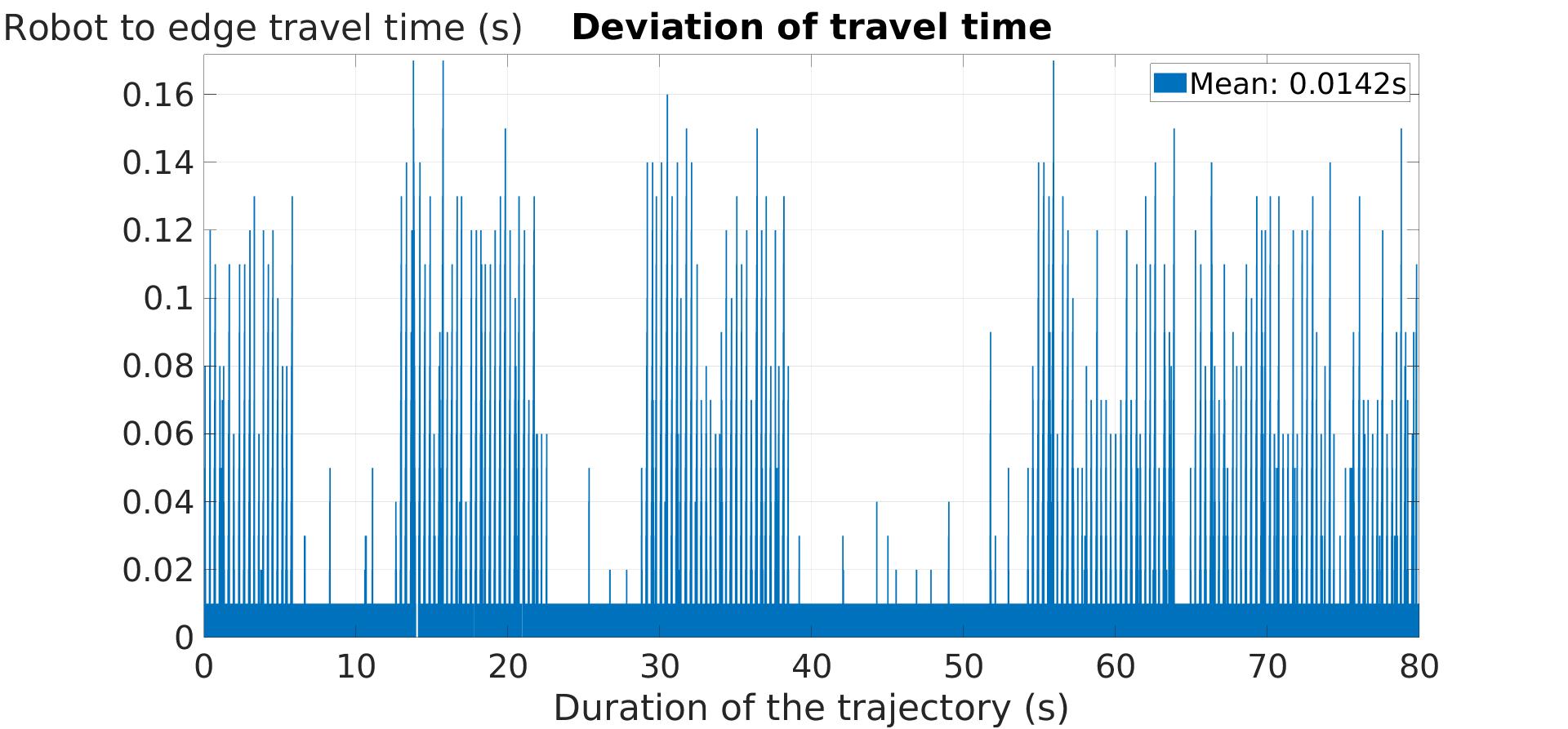}
  	\caption{Deviation of robot to edge travel time for the helical trajectory}
  	\label{fig:dhelicaltime1}
\end{figure}

\begin{figure}[ht!]
	\centering
	\includegraphics[width=0.95\columnwidth]{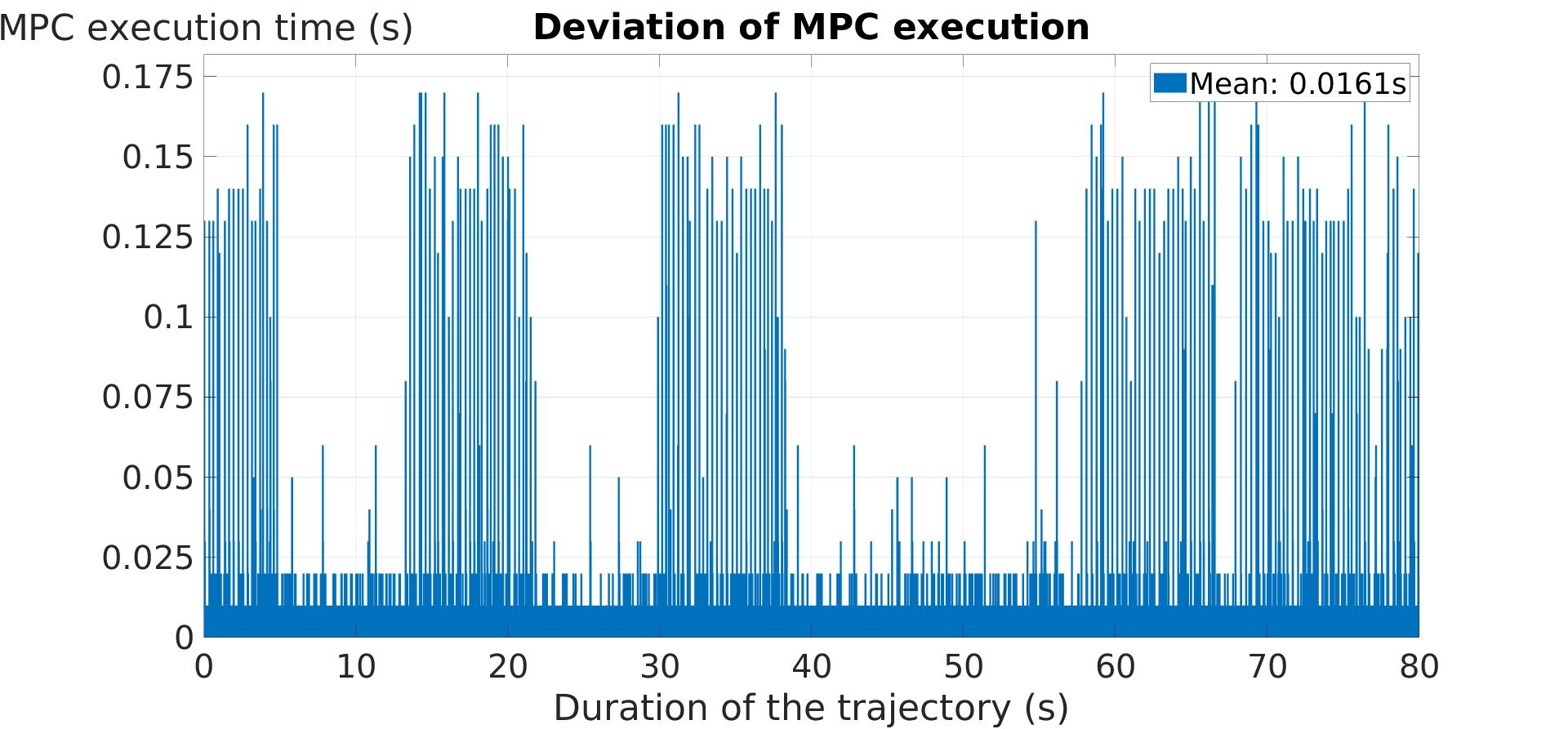}
  	\caption{Deviation of MPC execution time for the helical trajectory}
  	\label{fig:dhelicaltime2}
\end{figure}

\begin{figure}[ht!]
	\centering
	\includegraphics[width=0.95\columnwidth]{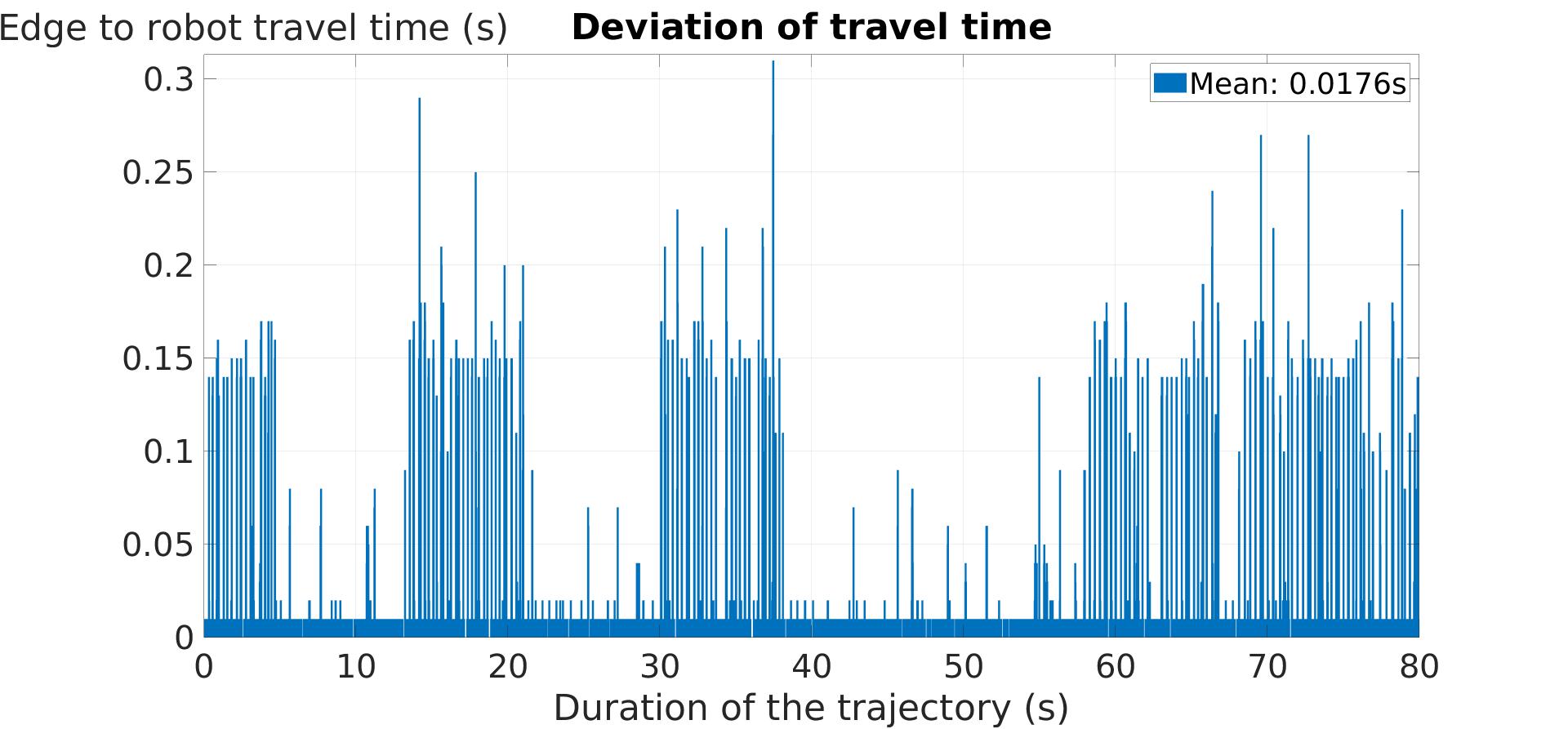}
  	\caption{Deviation of edge to robot travel time for the helical trajectory}
  	\label{fig:dhelicaltime3}
\end{figure}

The measurements from the helical trajectory for the travel and execution times are shown in In~Fig.~\ref{fig:dhelicaltime1}, Fig.~\ref{fig:dhelicaltime2} and Fig.~\ref{fig:dhelicaltime3}, where the duration of the trajectory was 80 seconds. The mean travel time of a packet from the robot to the edge is $14.2 milliseconds$, the execution time is $16.1 milliseconds$, 
and the travel time of a packet from the edge to the robot is $17.6 milliseconds$. We can observe some high values on the above graphs, which do not effect the response of the system, since they are not continuous and the values are not extremely high.

The overall time or round trip time (rtt) of the system depends on the MPC execution time, which is based on the MPC rate and computational needs, and the travel time delays, which are based on the network. The round trip time is the sum of all the each time parameter as shown in Eq.~\ref{eq:latencies}.

\begin{equation}
T_{rtt} = T_{ttre} + T_{exec} + T_{tter}
\label{eq:latencies}
\end{equation}

where $T_{rtt}$ is the round trip time, $T_{ttre}$ is the travel time from the robot to the edge, $T_{exec}$ is the MPC execution time, and $T_{tter}$ is the travel time from the edge to the robot. Thus, the measured mean round trip time is $47.9 milliseconds$.

\begin{figure}[ht!]
	\centering
	\includegraphics[width=0.95\columnwidth]{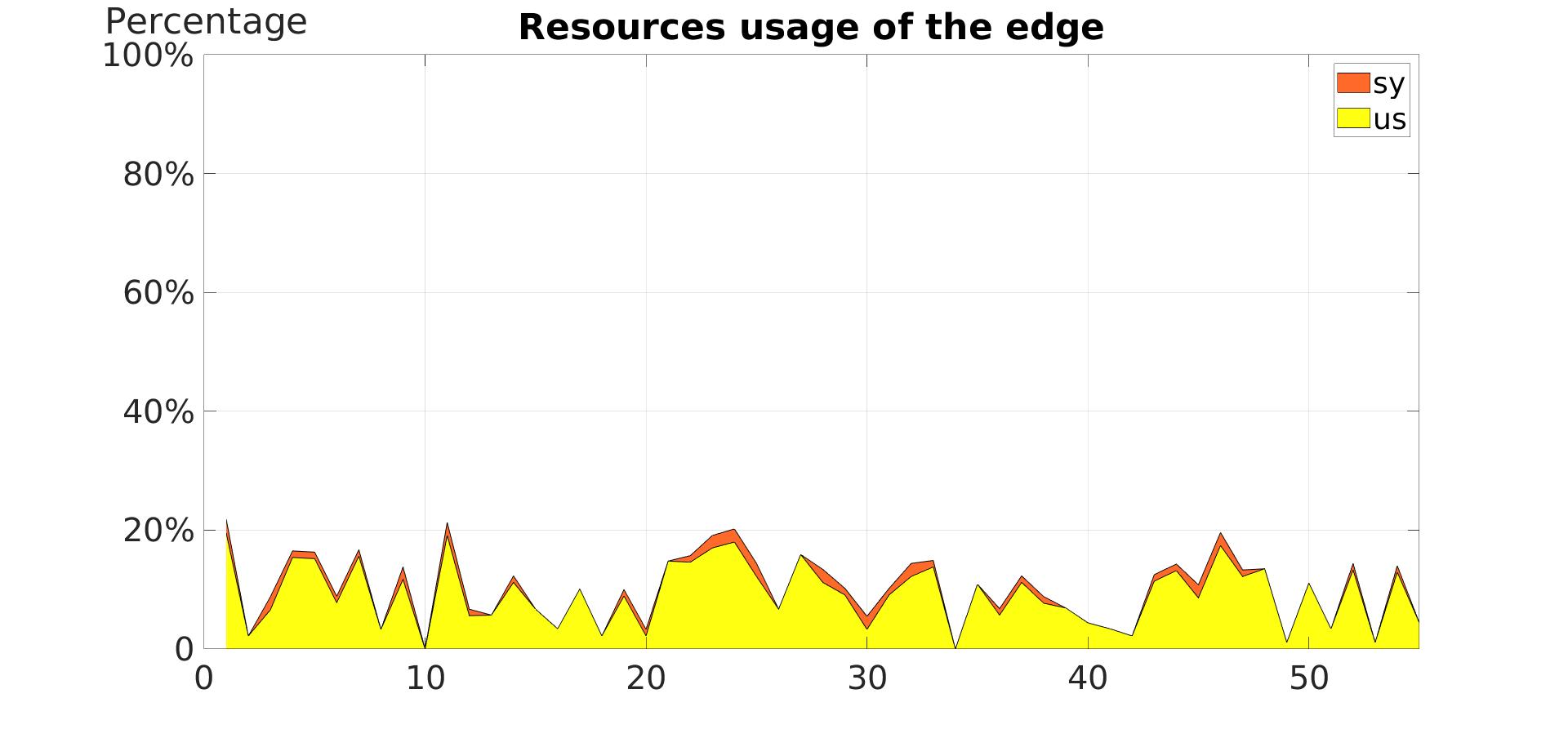}
  	\caption{Edge resources usage for the helical trajectory}
  	\label{fig:dhelicalcpu}
\end{figure}

The percentage of resources used at the edge for the execution of the controller is depicted in Fig.~\ref{fig:dhelicalcpu}. The yellow area represents the $us$ which is the amount of time the CPU spends executing processes in user-space, while the orange area represents $sy$ which is the amount of time spent running system kernel-space processes. The mean $us$ is $9.2000\%$ and the mean $sy$ is $0.8328\%$, while the mean of both processes combined is $10.0328\%$.


The next series of the simulation tests is based on the kubernetes architecture. 



\begin{figure}[ht!]
	\centering
	\includegraphics[width=0.95\columnwidth]{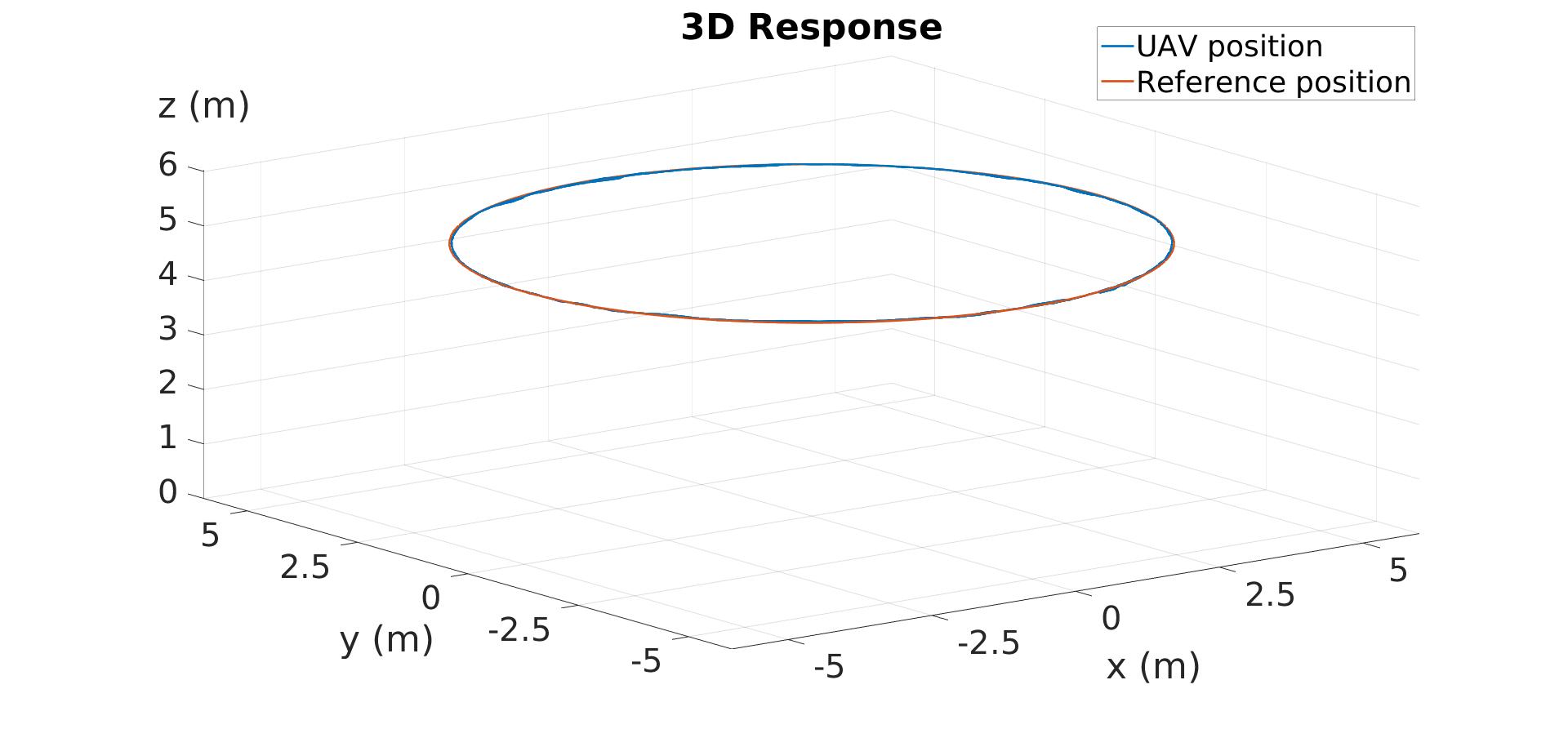}
  	\caption{UAV 3D circular trajectory based on a kubernetes-based architecture}
  	\label{fig:kcircular3d}
\end{figure}

\begin{figure}[ht!]
	\centering
	\includegraphics[width=0.95\columnwidth]{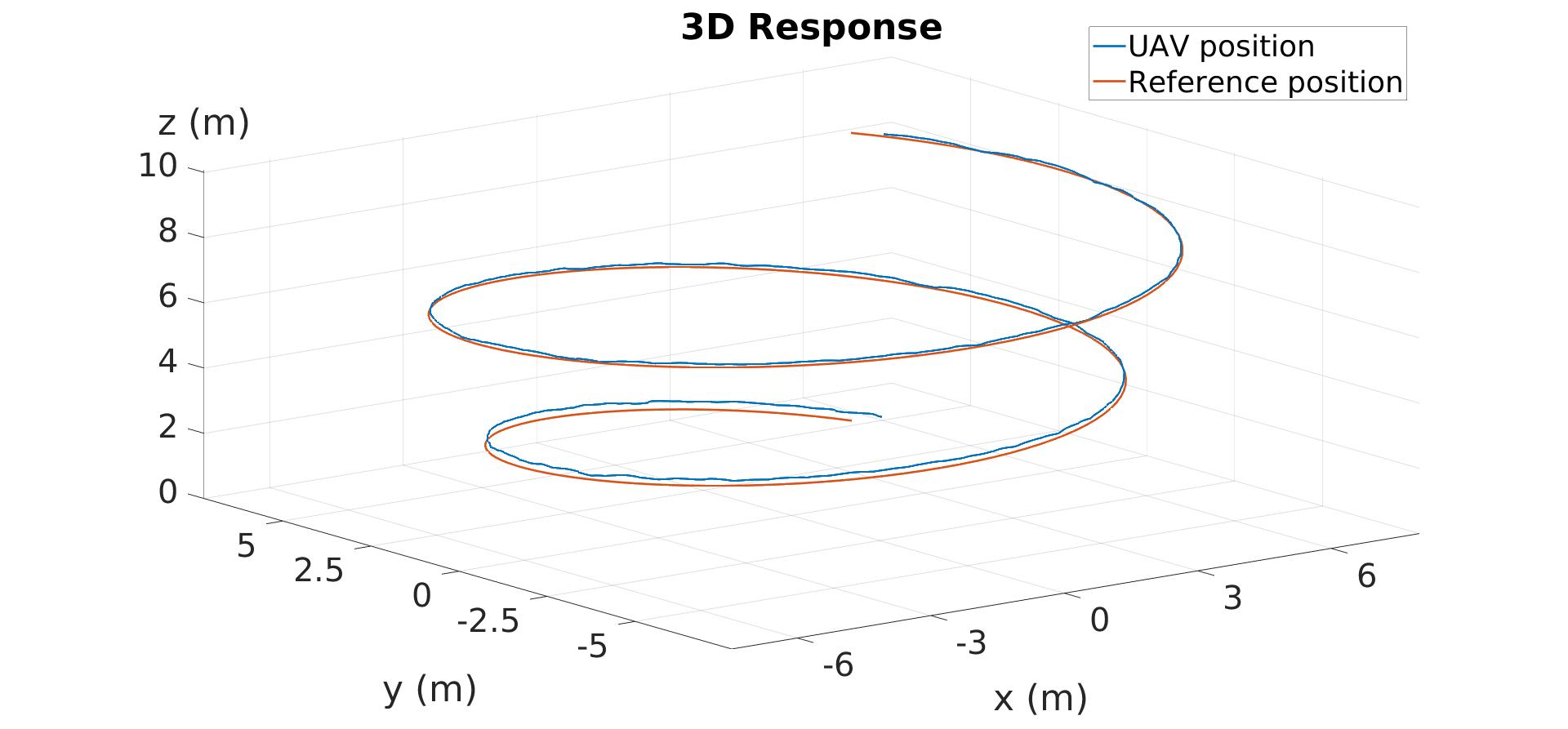}
  	\caption{UAV 3D helical trajectory based on a kubernetes-based architecture}
  	\label{fig:khelical3d}
\end{figure}

The responses are depicted in~Fig.~\ref{fig:kcircular3d} and Fig.~\ref{fig:khelical3d}. The closed loop system behavior is the desired one.

\begin{figure}[ht!]
	\centering
	\includegraphics[width=0.95\columnwidth]{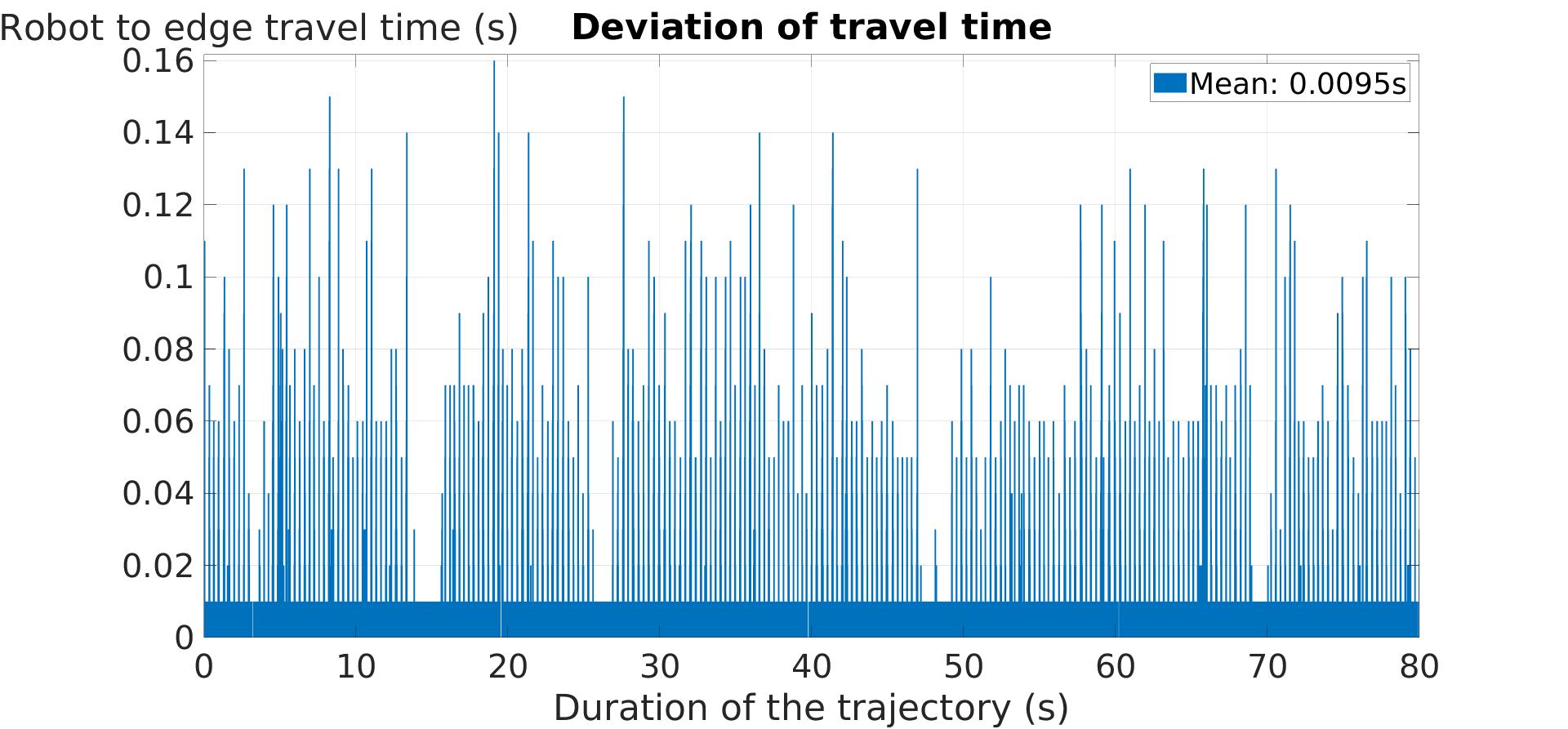}
  	\caption{Deviation of robot to edge travel time for the helical trajectory}
  	\label{fig:khelicaltime1}
\end{figure}

\begin{figure}[ht!]
	\centering
	\includegraphics[width=0.95\columnwidth]{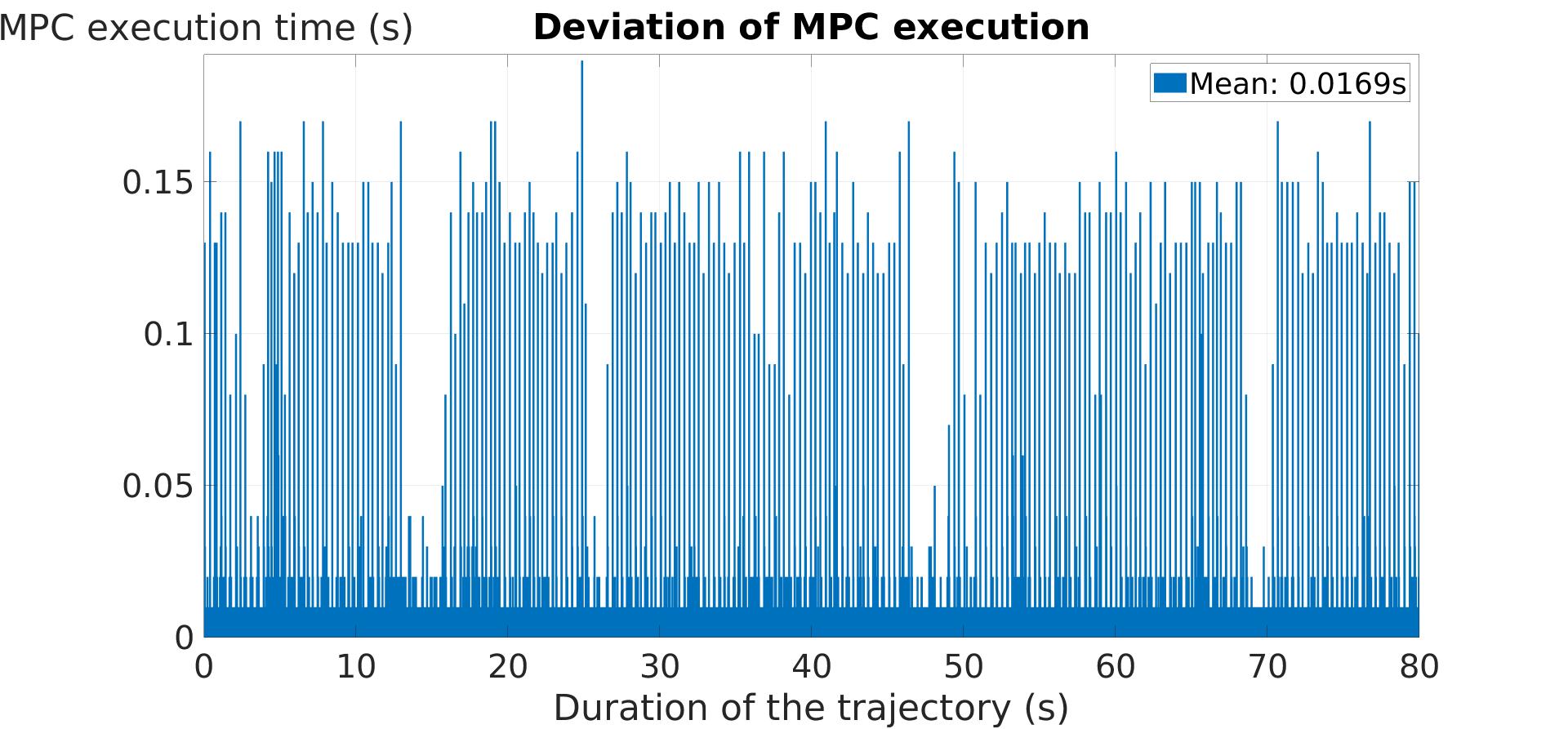}
  	\caption{Deviation of MPC execution time for the helical trajectory}
  	\label{fig:khelicaltime2}
\end{figure}

\begin{figure}[ht!]
	\centering
	\includegraphics[width=0.95\columnwidth]{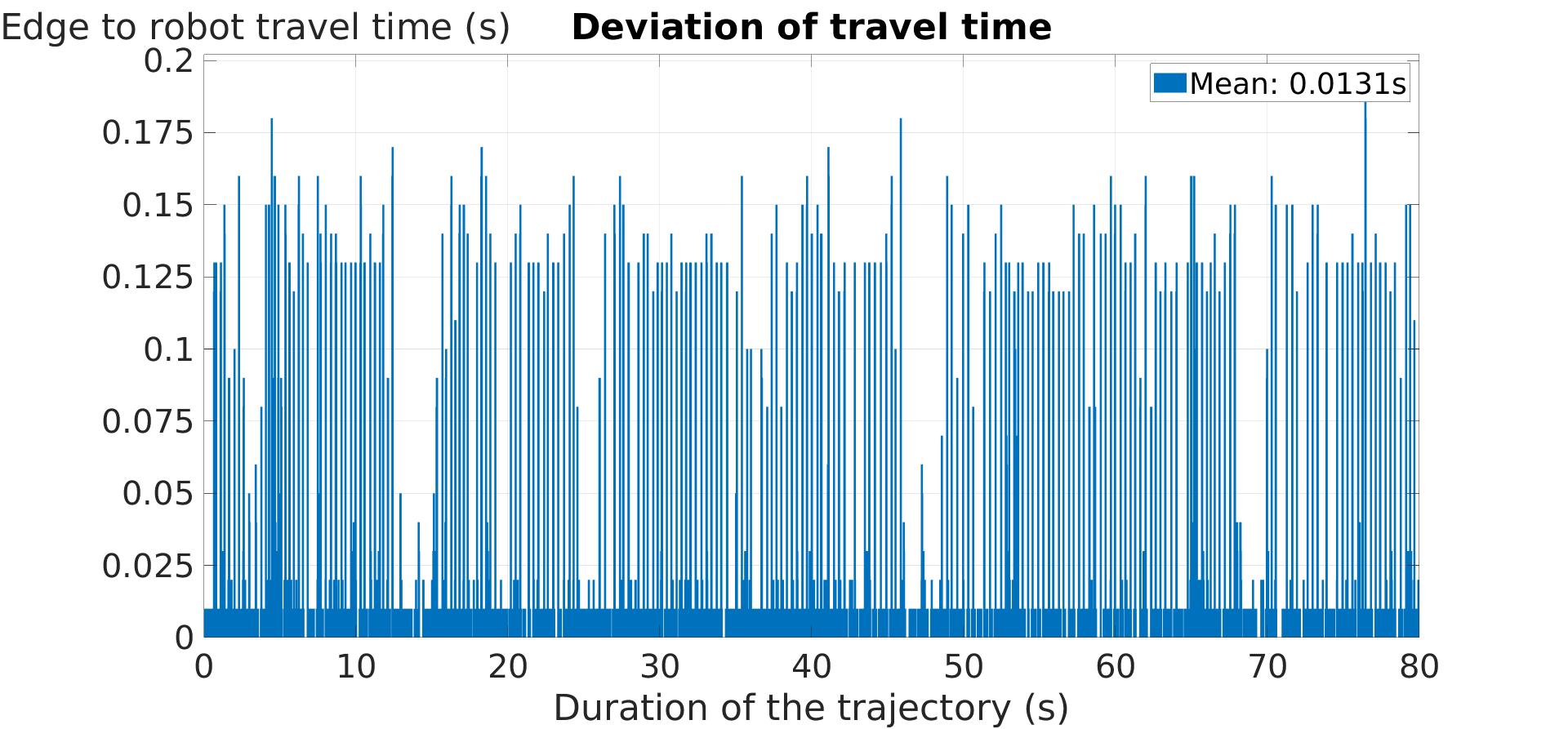}
  	\caption{Deviation of edge to robot travel time for the helical trajectory}
  	\label{fig:khelicaltime3}
\end{figure}

The results for the travel and execution times for the helical trajectory of the kubernetes-based architecture are shown in In~Fig.~\ref{fig:khelicaltime1}, Fig.~\ref{fig:khelicaltime2} and Fig.~\ref{fig:khelicaltime3}. The mean travel time of a packet from the robot to the edge is $9.5 milliseconds$, the execution time is $16.9 milliseconds$, and the travel time of a packet from the edge to the robot is $13.1 milliseconds$. Again we can notice some high values on the above graphs, which do not effect the response of the system.

The measured mean round trip time in this case is $39.5 milliseconds$, which is less than the measured round trip time of the docker architecture.

\begin{figure}[ht!]
	\centering
	\includegraphics[width=0.95\columnwidth]{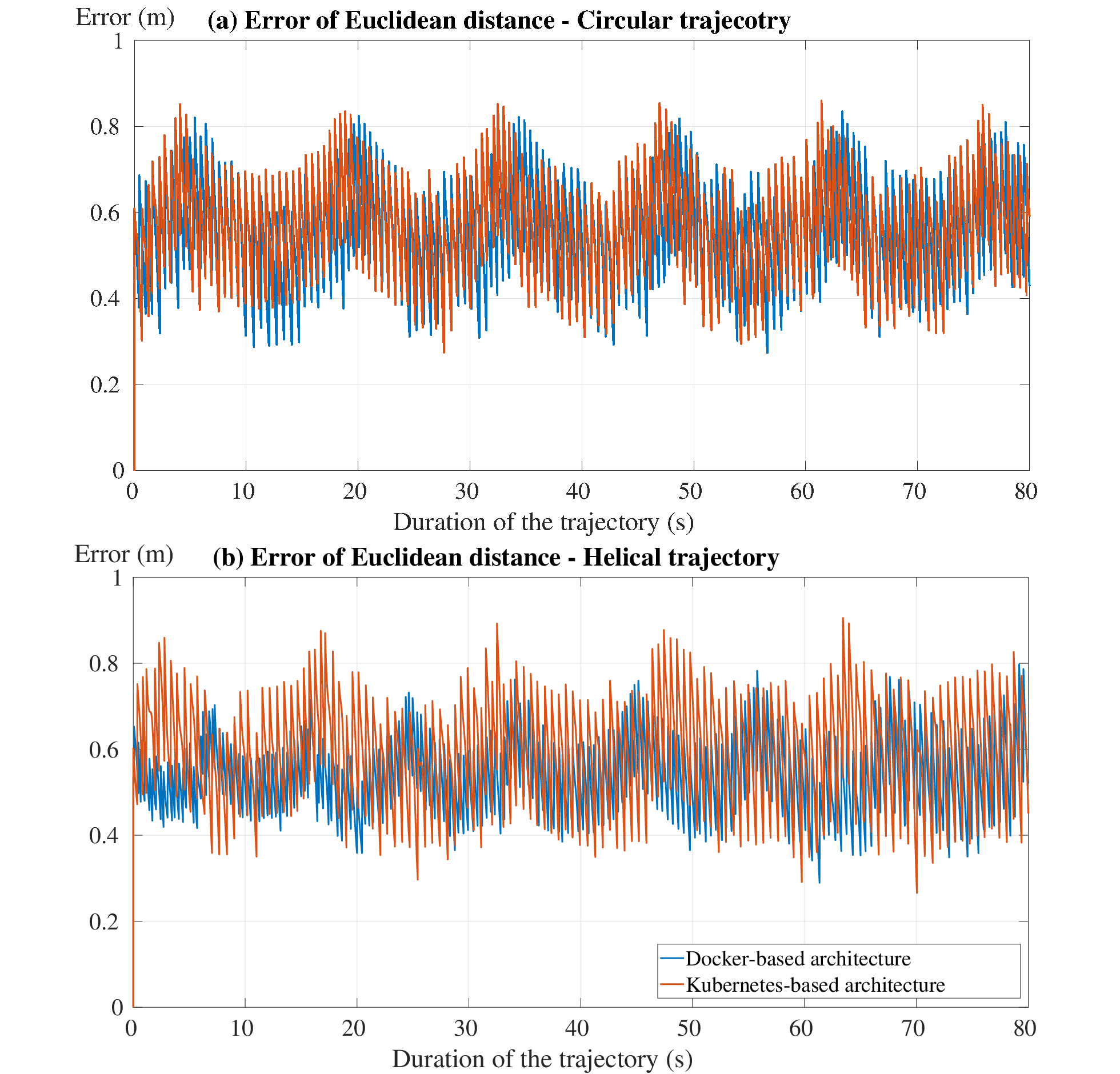}
  	\caption{Euclidean error between UAV position and reference position for docker-based (blue line) and kubernetes-based (red line) architectures, for (a) circular and (b) helical trajectory}
  	\label{fig:error}
\end{figure}

The Euclidean error between the actual position of the UAV and the reference position is shown in~\ref{fig:error}, where the blue line represents the error for the docker-based architecture and the red line represents the error for the kubernetes-based. In both cases. In this case the error is similar and is based on the chosen tolerance, which is set at $0.7 meters$.

\begin{figure}[ht!]
	\centering
	\includegraphics[width=0.95\columnwidth]{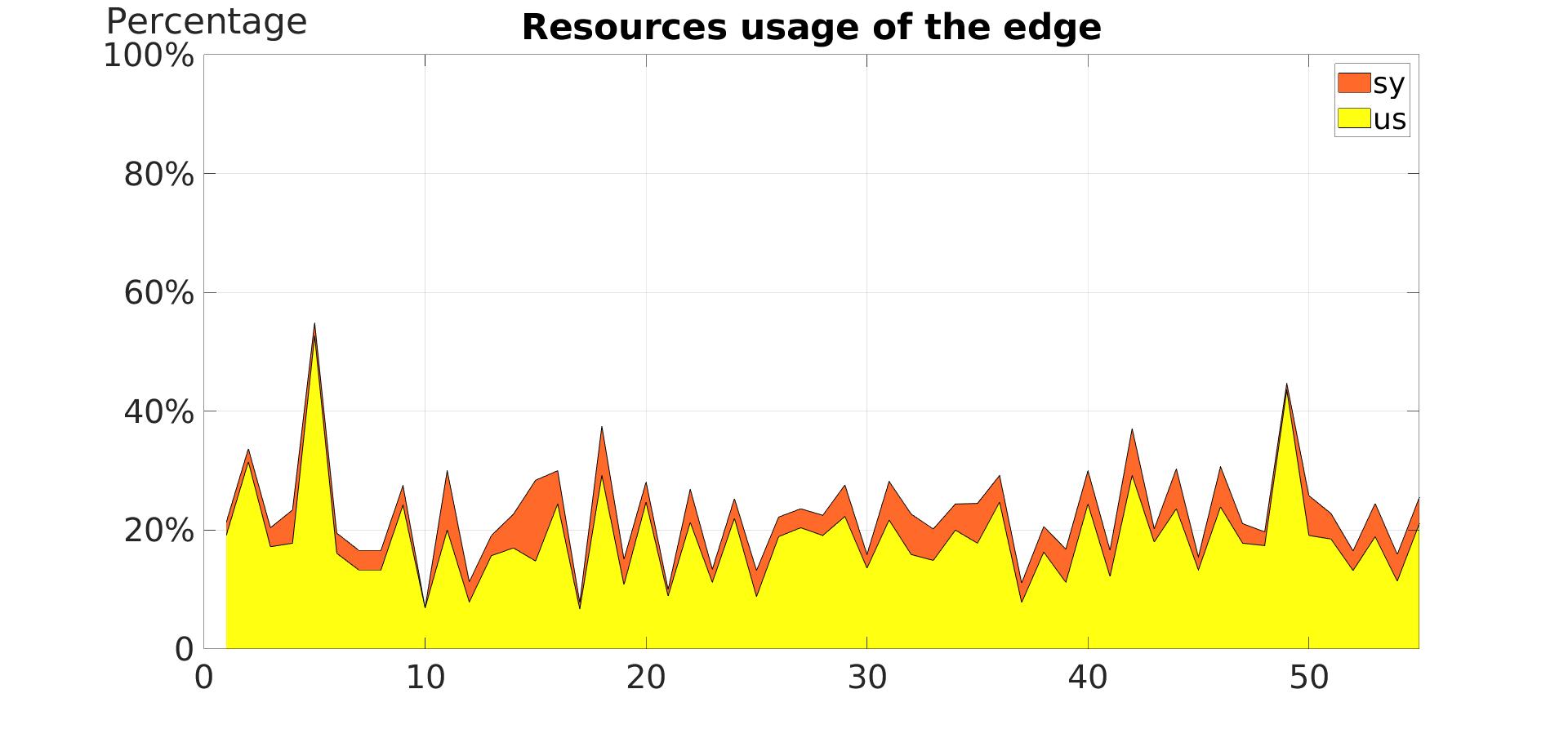}
  	\caption{Edge resources usage for the helical trajectory}
  	\label{fig:khelicalcpu}
\end{figure}

Finally, the percentage of resources used at the edge for the execution of the controller is depicted in Fig.~\ref{fig:khelicalcpu}. The mean $us$ is $18.8479\%$ and the mean $sy$ is $4.4603\%$, while the mean of both processes combined is $23.3082\%$. These values are quite bigger than the ones of the docker based architecture since the kubernetes environment is a more complicated and heavy than just a docker container, but still the edge machine can handle the process without any issues.

\section{Conclusions and Future Work}
\label{conclusion}
In this article, two novel edge architectures were presented and compared for controlling the trajectory of a UAV. We were able to successfully control the UAV with the implementation of both architectures, and the measured results were similar in both cases. Each of the architectures provides some advantages over the other one, depending on the application. A docker-based architecture is simple to implement but does not support some features that kubernetes architecture does that can be essential in some missions when container orchestration is needed. Based on the mission, we should be able to identify the best approach. Also the cost should be taken under consideration, since the use of a docker container without kubernetes orchestration is a much more affordable option. In this work, we were focused on containerized application forms and not on VMs, since the advantages that containers provide are essential for the kind of applications we were studying.

These architectures can be used in many scenarios for real-time operations that are time sensitive. The next step for these architectures would be, to be used in real-life scenarios in an experimental setup. Offloading the MPC to the edge is very useful since we were able to test a time sensitive system and set high values for the horizon and the execution rate to have desired performance. Other application that can be offloaded to the edge using the same architectures are the problem of simultaneous localization and mapping or path planning. The edge will not only provide resources but at the same time, it will provide a centralized layer through which robots will be able to communicate and collaborate with each other in order to complete missions autonomously.

\bibliographystyle{./IEEEtranBST/IEEEtran}
\bibliography{./IEEEtranBST/IEEEabrv,references}

\end{document}